\documentclass[letterpaper]{article} 
\usepackage{aaai23}  
\usepackage{times}  
\usepackage{helvet}  
\usepackage{courier}  
\usepackage[hyphens]{url}  
\usepackage{graphicx} 
\usepackage{natbib}  
\usepackage{caption} 
\usepackage{algorithm}
\usepackage{algorithmic}

\usepackage[table,xcdraw]{xcolor}
\usepackage{multirow}
\usepackage{amsthm,amsmath,amssymb}
\usepackage{subfigure}
\usepackage{etoolbox,lipsum}

\usepackage{newfloat}
\usepackage{listings}
\DeclareCaptionStyle{ruled}{labelfont=normalfont,labelsep=colon,strut=off} 
\floatstyle{ruled}
\newfloat{listing}{tb}{lst}{}
\floatname{listing}{Listing}
\title{GAN Prior based Null-Space Learning for Consistent Super-Resolution}
\author{
    Yinhuai~Wang\textsuperscript{\rm 1},
    Yujie~Hu\textsuperscript{\rm 1},
    Jiwen~Yu\textsuperscript{\rm 1},
    Jian~Zhang\textsuperscript{\rm 1,2}}
\affiliations{
    \textsuperscript{\rm 1}Peking University Shenzhen Graduate School, China\\
    \textsuperscript{\rm 2}Peng Cheng Laboratory, China

    \{yinhuai; hhuyujie; yujiwen\}@stu.pku.edu.cn, zhangjian.sz@pku.edu.cn
}

\usepackage{bibentry}

\begin{document}

\twocolumn[{
\renewcommand\twocolumn[1][]{#1}

\maketitle

\vspace{-0.2cm}
\begin{center}
    \centering
    \captionsetup{type=figure}
    \includegraphics[width=1\linewidth]{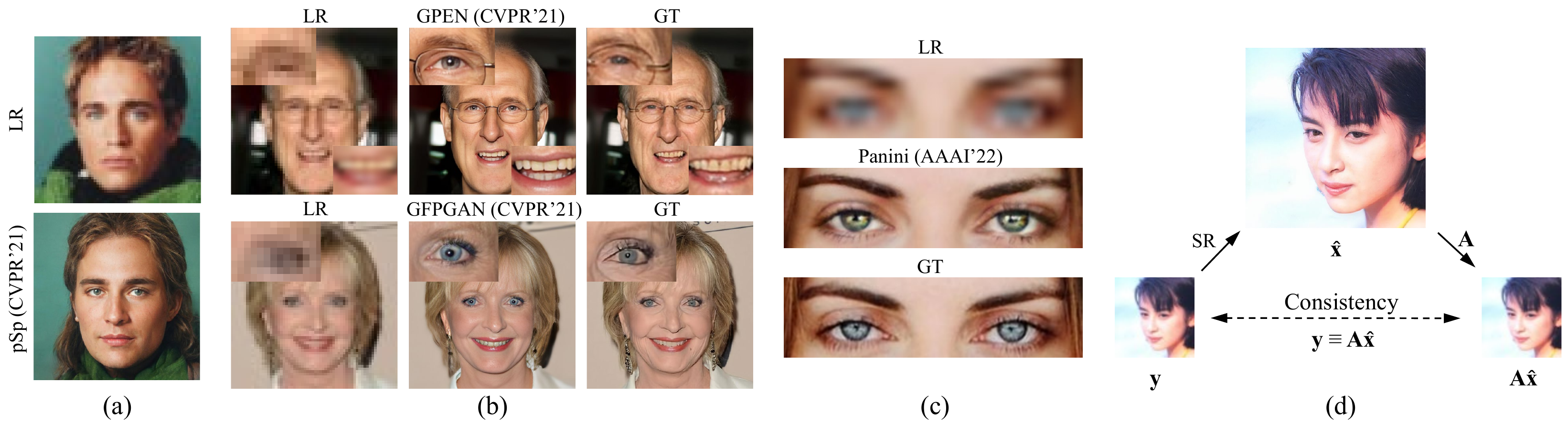}
    \captionof{figure}{\textbf{GAN prior based SR methods suffer inconsistencies.} Here we take four state-of-the-art GAN prior based SR methods for example. (a) pSp fails to maintain the identity of the low-resolution (LR) image; (b) GFP-GAN and GPEN can not faithfully restore the eye structures and colors of the LR; (c) Panini sometimes appears evident bias of eye colors during training. In these cases, it is clear that the LR images already contain rough structures and colors. However, the mentioned methods neglect these physical prior and entirely rely on networks to generate the SR result, causing inconsistencies. Instead, our proposed method assures downsampling consistency (defined by $\mathbf{A}\hat{\mathbf{x}}\equiv\mathbf{y}$) analytically, as is illustrated in (d), where $\mathbf{y}$, $\hat{\mathbf{x}}$, and $\mathbf{A}$ represents the LR image, SR result, and downsampling operator, respectively. Codes: \textcolor{cyan}{https://github.com/wyhuai/RND}.}
    \label{fig1} 
\end{center}
}]

\begin{abstract}
\textit{Consistency} and \textit{realness} have always been the two critical issues of image super-resolution. While the \textit{realness} has been dramatically improved with the use of GAN prior, the state-of-the-art methods still suffer inconsistencies in local structures and colors (e.g., tooth and eyes). In this paper, we show that these inconsistencies can be analytically eliminated by learning only the null-space component while fixing the range-space part. Further, we design a pooling-based decomposition (PD), a universal range-null space decomposition for super-resolution tasks, which is concise, fast, and parameter-free. PD can be easily applied to state-of-the-art GAN Prior based SR methods to eliminate their inconsistencies, neither compromise the \textit{realness} nor bring extra parameters or computational costs. Besides, our ablation studies reveal that PD can replace pixel-wise losses for training and achieve better generalization performance when facing unseen downsamplings or even real-world degradation. Experiments show that the use of PD refreshes state-of-the-art SR performance and speeds up the convergence of training up to \textbf{2}$\sim$\textbf{10} times.
\end{abstract}

\section{Introduction}

Image super-resolution (SR) is a process that generates a high-resolution (HR) image that is consistent with the input low-resolution (LR) image. It consists of two critical constraints: (1) Low-frequency consistency, i.e., the downsampling of the HR image should be identical to the LR image. (2) Realness, i.e., the HR image should be within the distribution of natural images. In this paper, we abbreviate these two constraints to \textit{consistency} and \textit{realness}.

Prevailing deep neural network (DNN) based SR methods usually relies on pixel-wise losses (e.g., $\ell_1$, $\ell_2$) to learn both \textit{realness} and \textit{consistency}. Specifically, they first downsample the ground-truth (GT) images to generate LR-GT image pairs, then feed the LR images into a DNN to generate an HR image and optimize the DNN by minimizing the pixel-wise losses between HR and GT. When the GT images are of high-quality, the pixel-wise losses measures both the \textit{realness} and \textit{consistency} constraints. Generative adversarial network (GAN) and the adversarial training objectives \cite{gan} are proved effective as a guidance toward \textit{realness} \cite{srgan,esrgan}. The Recent breakthrough in GANs enables generating high-resolution images \cite{stylegan,stylegan2,stylegan3}, and encourages utilizing pretrained GAN models as the GAN prior for SR tasks \cite{pulse, psp, gfpgan, gpen, glean, panini}.

Though superior in hallucinating realistic details, GAN prior may bring undesirable bias in generating those details, causing inconsistencies in structures and colors, e.g., eye, tooth, and skin, as can be seen in Fig.~\ref{fig1}. Further increasing the weight of $\ell_1$ or $\ell_2$ losses may ease these inconsistencies while over-smoothing the rich details that GAN Prior brings. Since \textit{realness} and \textit{consistency} are both critical for SR, we need to find a way to assure \textit{consistency} for GAN prior based SR without compromise its superiority in \textit{realness}.

We observe there exists a fundamental physical prior in SR tasks. That is, the LR image (assuming the LR image is clean) already contains low-frequency information that constitutes the GT image. In other words, the LR image already defined the rough color and structural distributions of the GT image. However, recent GAN prior based SR methods do not fully leverage this physical prior. To utilize this physical prior, we can resort to the range-null space decomposition, which is well studied in linear algebra and inverse problems \cite{dnl}.

Theoretically, given a downsampler and its pseudo-inverse, any HR image can be decomposed into a range-space part and a null-space part. The range-space part is fully responsible for the LR image while the null-space part is irrelevant to the LR image. Actually, the range-space part of the GT can be calculated explicitly. Then we can explicitly assure the \textit{consistency} of a predicted HR image by replacing its range-space part with the GT's range-space part.

However, the pseudo-inverse of the downsampler is usually hard to obtain, let alone so many types of downsamplers in SR tasks. We observe that most of the downsampling with antialiasing shares very similar results. We also notice a wildly used downsampling method, average-pooling, possesses concise forms of pseudo-inverse. Hence we use average-pooling as a universal measurement for the \textit{consistency}. This provides an efficient solution to use range-null space decomposition for consistent SR.

Specifically, we use the pseudo-inverse to directly upsample the LR image as the low-frequency part of the result, then use the GAN prior based network to predict a raw HR image from the LR image, and use the calculated null-space operator to extract its high-frequency part. Finally, we add the low-frequency part and the high-frequency part to form an SR result that owns the \textit{consistency}. The whole process is called Pooling-based Decomposition (PD) in this paper.

As a physical prior of \textit{consistency}, the involvement of PD not only enables existing state-of-the-art GAN prior based SR methods to reach obviously higher performance in \textit{consistency} but also significantly accelerates training convergence, as is revealed in our experiments. Besides, our ablation studies show that PD can replace pixel-wise losses for training and achieves better generalization performance when facing unseen downsamplings or real-world degradation, implying its potential for further explorations. 

Our contributions include:
\begin{itemize}
  \item We theoretically and experimentally reveal that combining GAN prior with physical prior yields significantly better \textit{consistency}, outperforming state-of-the-art GAN prior based SR methods in several datasets.
  \item We propose a novel Pooling-based Decomposition (PD) for GAN prior based null-space learning. It is parameter-free, fast, and mathematically rigorous. PD can be used for GAN prior based SR methods to improve the \textit{consistency} and accelerate training convergence significantly.
\end{itemize}

\section{Related Work}

\paragraph{GAN Prior based Super-Resolution.} 
Typical DNN based super-resolution (SR) methods use pixel-wise constraints to encourage learning the inverse mapping of downsampling \cite{srcnn, vdsr, rcan, rdn, wavelet, xiaoming, lin2018multi, lin2020learning}. These methods usually perform well in pixel-wise metrics, e.g., PSNR and SSIM, but tend to generate smooth structures with poor details. Owing to the excellent performance in generating realistic details, GANs and adversarial training strategies \cite{gan} are applied for SR to produce higher visual quality \cite{esrgan, xinyu, hifacegan, ziyu, menglei}. Recently, Karras et al. proposed a series of StyleGAN that can generate excellent realistic images of certain types, typically human faces \cite{stylegan,stylegan2,stylegan3}. The success of StyleGAN inspired a lot of works in image editing \cite{sam, styleflow, image2stylegan, image2stylegan++, iddisentanglement, stylerig, sean, styleclip, gu2020image, xia2022gan} and encouraged the utilizing of StyleGAN as GAN prior for SR tasks. Some methods attempt utilizing a pretrained GAN to solve SR by optimizing the latent code iteratively to maximize the data consistency \cite{Yeh,Luo,pulse}. Richardson et al. further propose using an encoder to map the input image into the latent space of StyleGAN \cite{psp}. Since they directly utilize the latent space of StyleGAN, the result owns excellent \textit{realness}. However, the latent space is not expressive enough to preserve the identity of the input image, thus suffering identity shifts. To improve the \textit{consistency}, some methods build skip connections between the encoder and StyleGAN to better capture structural information \cite{gpen, gfpgan,glean}. Wang et al. further propose interpolating StyleGAN features and encoder features to yield better \textit{consistency} and \textit{realness} \cite{panini}. These GAN prior based methods achieve unprecedented results in face SR. Though pixel-wise losses are already applied during their training, the use of GAN prior may still bring undesirable inconsistency in facial structures and colors, as is shown in Fig.~\ref{fig1}. Besides, the success of these methods relies heavily on trial and error, lacking rigorous explanation and mathematical proof. Instead, our proposed PD follows rigorous derivation and can be applied to the mentioned methods to completely resolve inconsistency while preserving the \textit{realness} that GAN prior brings.

\paragraph{Consistent Super-Resolution.}

Traditional approaches for solving super-resolution (SR) are typically model-based, which usually regularize the results with prior knowledge about distributions of natural images, e.g., sparsity \cite{sparsity}. 
Most of the recent state-of-the-art SR methods are DNN based and are model-free \cite{liang2021swinir,glean}, thus lack interpretation and heavily relying on extensive experiments on network structures. To implement a consistent solution for inverse problems, Chen et al. \cite{chen2020deep} analytically decompose the inverse result into range-space and null-space of a specific linear operator and learn them respectively. However, they take bicubic interpolation as the pseudo-inverse of bicubic interpolation for image SR, which is a lossy approximation. Bahat et al. \cite{cem} further resort to the Fourier domain to calculate the pseudo-inverse of the downsampler $\mathbf{A}$. However, their formulation contains inverse matrix $(\mathbf{A}\mathbf{A}^{\top})^{-1}$, which may not exist for not-full rank $\mathbf{A}\mathbf{A}^{\top}$. Hence the pseudo-inverse calculated by FFT and IFFT may still lossy (as is shown in Table \ref{table1}) in theory, let alone the complexity and inaccuracy in implementations. Besides, they did not combine null-space learning with GAN prior, thus can not yield high-quality details. Instead, we show that the GAN prior is an ideal tool to fill in the missing null-space in SR problems, and the implementations of our proposed PD perfectly fit the theory and possess concise forms.

\section{Method}

\subsection{Preliminaries: Range-Null Space Decomposition}

Given a non-zero linear operator $\mathbf{A}\in\mathbb{R}^{d\times D}$, it usually has at least one pseudo-inverse $\mathbf{A^{\dagger}}\in\mathbb{R}^{D\times d}$ that satisfies $\mathbf{A}\mathbf{A^{\dagger}\mathbf{A}}=\mathbf{A}$. In particular, an analytical solution by SVD: 
\begin{equation}
    \mathbf{A}=\mathbf{U}\mathbf{\Sigma}\mathbf{V}^{\top},\quad \mathbf{A^{\dagger}}=\mathbf{V}\mathbf{\Sigma}^{\dagger}\mathbf{U}^{\top},
    \label{eq:1}
\end{equation}
where $\mathbf{U}$ and $\mathbf{V}$ are orthogonal matrix and $\mathbf{\Sigma}$ is a diagonal matrix with eigenvalues as its diagonal elements. $\mathbf{A^{\dagger}}\mathbf{A}$ can be seen as the operator that projects samples to the range-space of $\mathbf{A}$, since $\mathbf{A}\mathbf{A^{\dagger}\mathbf{A}}\equiv\mathbf{A}$. While $(\mathbf{I} - \mathbf{A^{\dagger}}\mathbf{A})$ can be seen as the operator that projects samples to the null-space of $\mathbf{A}$, since $\mathbf{A}(\mathbf{I} - \mathbf{A^{\dagger}}\mathbf{A})\equiv\mathbf{0}$.

Any sample $\mathbf{x}\in\mathbb{R}^{D\times 1}$ can be decomposed into two parts: the part that locates at the range-space of $\mathbf{A}$ and the other part that locates at the null-space of $\mathbf{A}$, i.e.,
\begin{equation}
    \mathbf{x}\equiv\mathbf{A^{\dagger}}\mathbf{A}\mathbf{x} + (\mathbf{I} - \mathbf{A^{\dagger}}\mathbf{A})\mathbf{x}
    \label{eq:2}
\end{equation}

\subsection{GAN Prior based Null-Space Learning}
In image super-resolution, the linear operator $\mathbf{A}$ becomes a downsampler while its pseudo-inverse $\mathbf{A}^{\dagger}$ represents an upsampler. We consider naive downsampling of the form:
\begin{equation}
    \mathbf{y}= \mathbf{A}\mathbf{x}.
    \label{eq:3}
\end{equation}
Given a low-resolution (LR) image $\mathbf{y}\in\mathbb{R}^{d\times 1}$ that downsampled from a ground-truth (GT) image $\mathbf{x}\in\mathbb{R}^{D\times 1}$ using downsampling operator $\mathbf{A}\in\mathbb{R}^{d\times D}$, our goal is to get a high-resolution (HR) image $\hat{\mathbf{x}}\in\mathbb{R}^{D\times 1}$ that conforms to:
\begin{equation}
    Consistency: \quad \mathbf{A}\hat{\mathbf{x}} \equiv \mathbf{y},
    \label{eq:4}
\end{equation}
\begin{equation}
    Realness: \quad \hat{\mathbf{x}} \sim p(\mathbf{x}),
    \label{eq:5}
\end{equation}
where $p(\mathbf{x})$ denotes the GT distribution. 
Though these two properties can both be optimized by training a DNN using pixel-wise and adversarial objectives, it does not thoroughly utilize the physical prior contained in the LR image.  

We observe that the range-null space decomposition is an ideal tool to ensure \textit{consistency}. Lets first decompose the GT $\mathbf{x}$ into two parts follow Eq.~(\ref{eq:2}), then downsample $\mathbf{x}$ with $\mathbf{A}$, i.e., combining Eq.~(\ref{eq:2}) with Eq.~(\ref{eq:3}), it becomes:
\begin{equation}
    \begin{aligned}
        \mathbf{A}\mathbf{x}&=\mathbf{A}\mathbf{A^{\dagger}}\mathbf{A}\mathbf{x} + \mathbf{A}(\mathbf{I} - \mathbf{A^{\dagger}}\mathbf{A})\mathbf{x}\\
        &= \mathbf{A}\mathbf{x} + \mathbf{0} = \mathbf{y}.
        \label{eq:6}
    \end{aligned}
\end{equation}
We can see that the range-space part, $\mathbf{A^{\dagger}}\mathbf{A}\mathbf{x}$, after downsampling, becomes exactly the LR image $\mathbf{y}$, while the null-space part, $(\mathbf{I} - \mathbf{A^{\dagger}}\mathbf{A})\mathbf{x}$, is transparent to the downsampler $\mathbf{A}$. Following these observations, we can formulate our HR result $\hat{\mathbf{x}}$ into two parts: a range-space part that is set as the GT's range-space part $\mathbf{A}^{\dagger}\mathbf{A}\mathbf{x}$, i.e., $\mathbf{A}^{\dagger}\mathbf{y}$; a null-space part $(\mathbf{I}-\mathbf{A}^{\dagger}\mathbf{A})\hat{\mathbf{x}}_{r}$ extracted from the raw prediction $\hat{\mathbf{x}}_{r}$ of a GAN prior based network.
\begin{equation}
    \hat{\mathbf{x}} = \mathbf{A}^{\dagger}\mathbf{y} + (\mathbf{I}-\mathbf{A}^{\dagger}\mathbf{A})\hat{\mathbf{x}}_{r},
    \label{eq:7}
\end{equation}
Now the result $\hat{\mathbf{x}}$ owns \textit{consistency} since $\mathbf{A}\hat{\mathbf{x}} \equiv \mathbf{y}$ holds. 

However, it is usually hard to get the pseudo-inverse of $\mathbf{A}$, let alone the situation that $\mathbf{A}$ is unknown.

\subsection{Pooling-based Decomposition}
We observe that many downsampling methods with antialiasing share very similar results.
Among them, average-pooling is the most simple one and owns desirable forms of pseudo-inverse. Since we focus on the structural and color inconsistencies in GAN prior based SR methods, 
the downsampling consistency measured by average-pooling is robust enough to assure that.

Specifically, for a certain downsampling rate $s$, average-pooling split the HR image $\mathbf{x}$ into patches of size $s\times s$, and calculate each patch's average value as the value of a corresponding pixel in the LR image $\mathbf{y}$. A simple pseudo-inverse of average-pooling replicates each pixel of the LR image into corresponding patches of size $s\times s$. 

Considering operation on a single pixel, the average-pooling with scale $s$ can be represented as a $s^{2}\times 1$ matrix $\mathbf{A}$ with its pseudo-inverse as a $1\times s^{2}$ matrix $\mathbf{A}^{\dagger}$:
\begin{equation}
    \mathbf{A}^{\dagger} = (1, ..., 1)^{\top},\quad \mathbf{A} = (1/s^{2}, ..., 1/s^{2}).
    \label{eq:8}
\end{equation}

It is obvious that $\mathbf{A}\mathbf{A}^{\dagger}=\mathbf{I}$ holds. Since the operations between each pixel or each patch are independent, this formulation can be easily promoted to the whole image.

In practice, we use average-pooling in linear operator form, denoted as $\mathcal{P}_{\downarrow}(\cdot)$, and use broadcast mechanism to implement the pseudo-inverse of average-pooling, denoted as $\mathcal{P}_{\uparrow}(\cdot)$. Note that $\mathcal{P}_{\downarrow}(\mathcal{P}_{\uparrow}(\cdot))=\mathcal{I}(\cdot)$, where $\mathcal{I}(\cdot)$ is the unit linear operator. Then we can rewrite our SR solution as:
\begin{equation}
    \hat{\mathbf{x}} = \mathcal{P}_{\uparrow}(\mathbf{y}) + \hat{\mathbf{x}}_{r} - \mathcal{P}_{\uparrow}(\mathcal{P}_{\downarrow}(\hat{\mathbf{x}}_{r})),
    \label{eq:9}
\end{equation}
where we use a GAN prior based DNN $\mathcal{D}_{\uparrow}$ to predict $\hat{\mathbf{x}}_{r}$:
\begin{equation}
    \hat{\mathbf{x}}_{r} = \mathcal{D}_{\uparrow}(\mathbf{y}),
    \label{eq:10}
\end{equation}
Note Eq.~(\ref{eq:9}) satisfies the downsampling consistency:
\begin{equation}
\begin{aligned}
    \mathcal{P}_{\downarrow}(\hat{\mathbf{x}}) &=  \mathcal{P}_{\downarrow}(\mathcal{P}_{\uparrow}(\mathbf{y})) + \mathcal{P}_{\downarrow}(\hat{\mathbf{x}}_{r}) - \mathcal{P}_{\downarrow}(\mathcal{P}_{\uparrow}(\mathcal{P}_{\downarrow}(\hat{\mathbf{x}}_{r}))),\\
    &= \mathbf{y} + \mathcal{P}_{\downarrow}(\hat{\mathbf{x}}_{r}) - \mathcal{P}_{\downarrow}(\hat{\mathbf{x}}_{r})\equiv\mathbf{y}.
\end{aligned}
\label{eq:11}
\end{equation}

\begin{figure*}[t]
  \centering
  \includegraphics[width=1\linewidth]{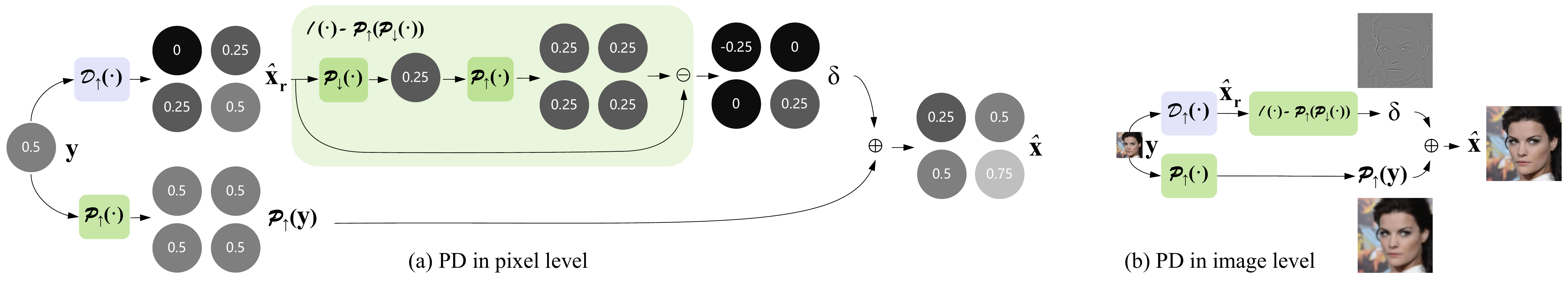}
  \caption{Illustration of pooling-based decomposition (PD). $\mathcal{D}_{\uparrow}(\cdot)$ denotes the network. $\mathcal{P}_{\downarrow}(\cdot)$ and $\mathcal{P}_{\uparrow}(\cdot)$ denote the average pooling and its pseudo-inverse. (a) shows a pixel level example of PD in 2$\times$ SR. For each pixel in the LR image $\mathbf{y}$, we use $\mathcal{P}_{\uparrow}(\cdot)$ to replicate it to the size of 2$\times$2. For the corresponding patch that predicted by the network $\mathcal{D}_{\uparrow}$, we subtract its own average value to get the high-frequencies $\hat{\mathbf{x}}_{r} - \mathcal{P}_{\uparrow}(\mathcal{P}_{\downarrow}(\hat{\mathbf{x}}_{r}))$, denoted as $\mathbf{\delta}$. Then we aggregate $\mathbf{\delta}$ and $\mathcal{P}_{\uparrow}(\mathbf{y})$ as the result patch $\mathbf{\hat{x}}$. Note $\mathbf{\hat{x}}$ is consistent with $\mathbf{y}$, i.e., $\mathcal{P}_{\downarrow}(\mathbf{\hat{x}}) \equiv \mathbf{y}$. We can easily promote this paradigm to the image level, as is shown in (b).}
\label{fig2} 
\end{figure*}

We name the whole operations of Eq.~(\ref{eq:9}) as Pooling-based Decomposition, abbreviated as PD in this paper. Fig.~\ref{fig2} provides a detailed illustration of PD at pixel and image levels. $\mathcal{D}_{\uparrow}(\cdot)$ denotes the GAN prior based SR network. $\mathcal{P}_{\downarrow}(\cdot)$ and $\mathcal{P}_{\uparrow}(\cdot)$ denote the average pooling and its pseudo-inverse. Fig.~\ref{fig2}(a) shows a pixel level example of PD in 2$\times$ SR. For each pixel in the LR image $\mathbf{y}$, we use $\mathcal{P}_{\uparrow}(\cdot)$ to replicate it to the size of 2$\times$2. For the corresponding patch $\hat{\mathbf{x}}_{r}$ that predicted by the network $\mathcal{D}_{\uparrow}(\cdot)$, we subtract $\hat{\mathbf{x}}_{r}$ by its low-frequency contents $\mathcal{P}_{\uparrow}(\mathcal{P}_{\downarrow}(\hat{\mathbf{x}}_{r}))$ to get the high-frequencies $\mathbf{\delta}$, i.e., $\mathbf{\delta} = \hat{\mathbf{x}}_{r} - \mathcal{P}_{\uparrow}(\mathcal{P}_{\downarrow}(\hat{\mathbf{x}}_{r}))$. Then we aggregate $\mathbf{\delta}$ and $\mathcal{P}_{\uparrow}(\mathbf{y})$ as the result patch $\mathbf{\hat{x}}$. It is worth noting that $\mathbf{\hat{x}}$ is consistent with $\mathbf{y}$, i.e., $\mathcal{P}_{\downarrow}(\mathbf{\hat{x}}) \equiv \mathbf{y}$. We can easily promote this paradigm to the image level, as is shown in Fig.~\ref{fig2}(b).

Given a GAN prior based SR network, we can significantly elevate its \textit{consistency} by simply imposing PD to its forward pipeline following Eq.~(\ref{eq:9}) and Eq.~(\ref{eq:10}) and applying the new forward pipeline to the training and inference. It is worth mentioning that PD is parameter-free and incurs negligible extra computations, making it an efficient tool for GAN Prior based SR networks to eliminate inconsistencies.

PD can be understood in many ways. Intuitively, $\mathcal{P}_{\uparrow}(\mathbf{y})$ can be seen as the low-frequency part (upsampled to match the size of $\mathbf{x}_{r}$) that is directly inhereted from the LR $\mathbf{y}$ and $\hat{\mathbf{x}}_{r} - \mathcal{P}_{\uparrow}(\mathcal{P}_{\downarrow}(\hat{\mathbf{x}}_{r}))$ be the high-frequency part that is extracted from the network prediction $\hat{\mathbf{x}}_{r})$. The operation $\mathcal{P}_{\downarrow}(\cdot)$ can be seen as a low-frequency filter, and the high-frequency part $\hat{\mathbf{x}}_{r} - \mathcal{P}_{\uparrow}(\mathcal{P}_{\downarrow}(\hat{\mathbf{x}}_{r}))$ yields no information after filtered by $\mathcal{P}_{\downarrow}(\cdot)$, since $\mathcal{P}_{\downarrow}(\hat{\mathbf{x}}_{r} - \mathcal{P}_{\uparrow}(\mathcal{P}_{\downarrow}(\hat{\mathbf{x}}_{r})))\equiv\mathbf{0}$. Alternatively, we can see $\mathcal{P}_{\uparrow}(\mathbf{y}) - \mathcal{P}_{\uparrow}(\mathcal{P}_{\downarrow}(\hat{\mathbf{x}}_{r}))$ as the correction for the low-frequency contents of the initial prediction $\hat{\mathbf{x}}_{r}$. Or mathematically, we can see $\mathcal{P}_{\uparrow}(\mathbf{y})$ as the range-space part, which losses no information through observation $\mathcal{P}_{\downarrow}(\cdot)$, hence is set as fixed to assure consistency. While $\hat{\mathbf{x}}_{r} - \mathcal{P}_{\uparrow}(\mathcal{P}_{\downarrow}(\hat{\mathbf{x}}_{r}))$ is the null-space part, which leaves no information through observation $\mathcal{P}_{\downarrow}(\cdot)$, hence has fully flexibility for learning.

\section{Experiments}
We validate PD on two typical GAN prior based SR networks: Panini \cite{panini} and GLEAN \cite{glean}. We experiment SR on three typical categories: human face, cat, and church.

\begin{table}[t]
    \centering
    \begin{tabular}{lccc}
      \hline
         \rule{0pt}{10pt}{{Method}} & PSNR(LR)↑ & Time↓ \\
         \hline
         \rule{0pt}{10pt}{CEM} &42.2 &31.8ms\\
         \rule{0pt}{10pt}{PD} &\textbf{145.7} &\textbf{0.68ms}\\
         
         \hline
    \end{tabular}
    \caption{\textbf{Validation of the \textit{consistency}}. To compare PD with CEM, we calculate the \textit{consistency} strictly following their theories, respectively. The result shows that the implementation of PD is faster and more precise.}
    \label{table1}
\end{table}

\subsection{Validation of Theory}
We theoretically proved that PD can inherently assure the low-frequency consistency: $\mathcal{P}_{\downarrow}(\mathbf{\hat{x}})\equiv\mathbf{y}$. To verify that, we take an untrained Panini as the backbone (i.e., Panini with randomly initialized parameters) and use PD to generate the HR results. We take 100 images from the CelebA-HQ dataset and use 8$\times$ average-pooling to generate the LR images. We take the LR image as the input $\mathbf{y}$ and use the ``Panini w/ PD" to generate HR $\hat{\mathbf{x}}$, then calculate the PSNR($\mathbf{y}$, $\mathcal{P}_{\downarrow}(\mathbf{\hat{x}})$). The average PSNR of 100 image pairs reaches \textbf{145.7}, indicating that $\mathbf{y} \equiv \mathcal{P}_{\downarrow}(\mathbf{\hat{x}})$ holds with the use of PD. We also calculate the consistency of CEM, which use the Fourier transforms to calculate the pseudo-inverse $\mathbf{A^{\dagger}}$ of the 8$\times$ bicubic(antialias) downsampler $\mathbf{A}$. However, the average PSNR($\mathbf{y}$, $\mathbf{A^{\dagger}}\mathbf{\hat{x}}$) is only \textbf{42.2}, indicating that the implementations of CEM suffer distortion. See Tab.~\ref{table1}.

We also compare the inference speed of PD and CEM. Specifically, we use random noise as $\mathbf{\hat{x}_{r}}$ to exclude the use of networks, then use PD and CEM, respectively. We calculate the average running time of 100 inferences on a single Nvidia Terga4 GPU. We get 0.68ms for PD and 31.8ms for CEM, implying that PD is more efficient in execution.

\begin{figure*}[t]
  \centering
  \includegraphics[width=1\linewidth]{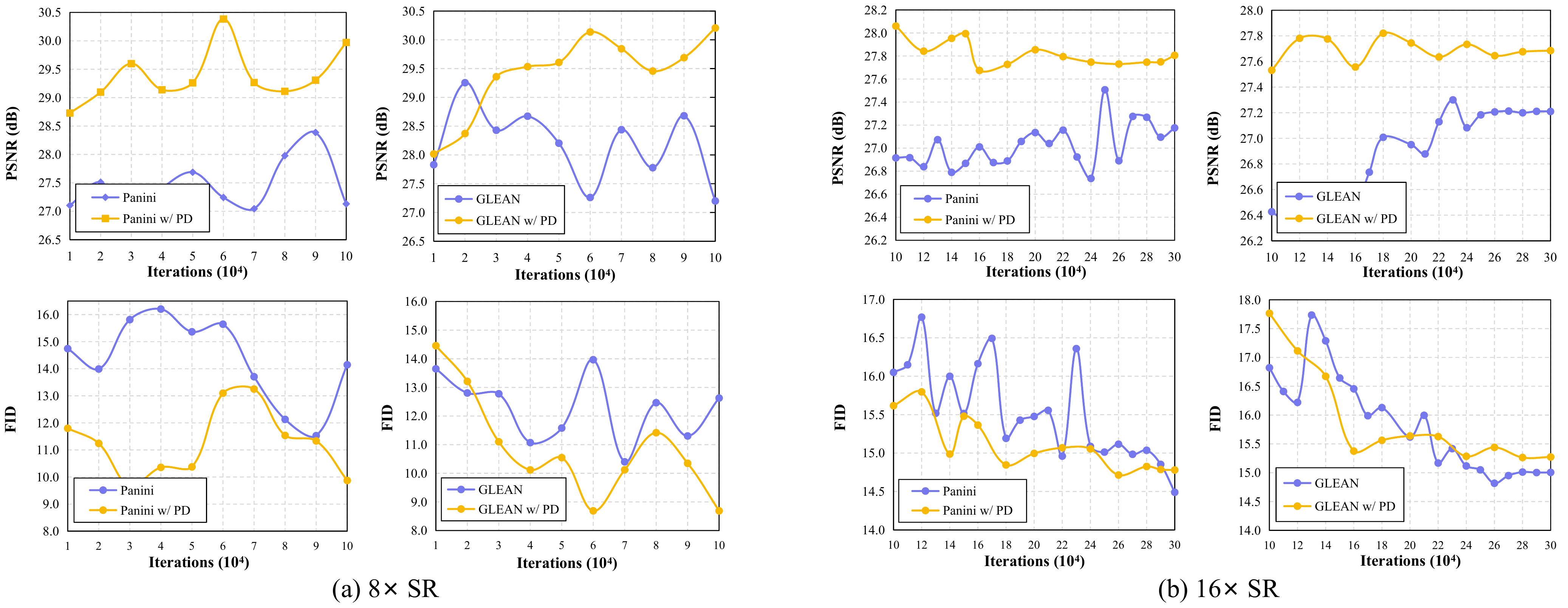}
  \caption{\textbf{Convergence curves}. Part (a) for the 8$\times$ face SR and (b) for the 16$\times$ face SR. With PD, both GLEAN and Panini yield significantly higher PSNR and comparable FID.}
\label{curve} 
\end{figure*}

\begin{figure}[t]
  \centering
  \includegraphics[width=1\linewidth]{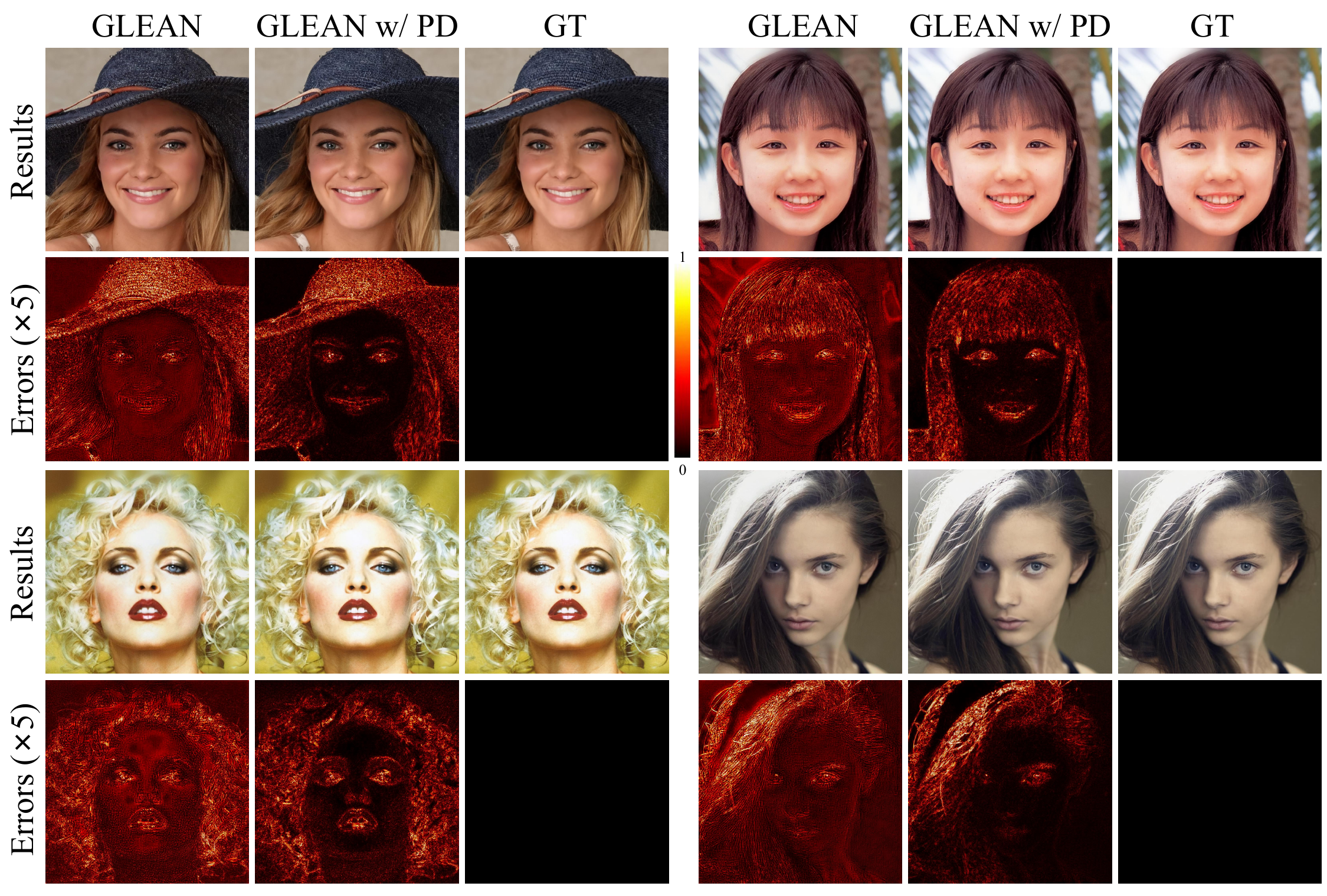}
  \caption{\textbf{Qualitative results on 8$\times$ face SR}. GLEAN yields high-quality results. However, it still suffers low-frequency inconsistencies, e.g., inconsistent structures and colors in lips, eyes, and large areas of skin color deviations. The use of PD helps eliminate all the low-frequency inconsistencies, as can be clearly observed from the error maps.}
\label{fig_8xsr} 
\end{figure}

\subsection{8$\times$ Super-Resolution} 
To experiment 8$\times$ SR on the human face, we train Panini, GLEAN, and their PD-based version on the FFHQ dataset \cite{stylegan}.
We use bicubic interpolation to synthesize LR images and follow the same training configuration as GLEAN and Panini. In detail, we use the Adam optimizer and Cosine Annealing Scheme with three training objectives: $\ell_1$ loss, perceptual loss \cite{perceptualloss}, and GAN loss \cite{gan}. For 8$\times$ SR, the loss weights are set as 1, 1$\times10^{-2}$, 1$\times10^{-2}$ respectively. The learning rate is set as 1$\times10^{-3}$. See Appendix A for more details of training objectives.

For 8$\times$ SR, we train Panini, ``Panini w/ PD", GLEAN, and ``GLEAN w/ PD" under the same training configuration for 100K iterations, with the batch size of 4 on a single Nvidia V100 GPU. For evaluation, we take 1K images from CelebA-HQ \cite{stylegan} dataset as the ground truth (GT), then use bicubic interpolation to generate the LR and yield the LR-GT testing pairs. 

\begin{table}[t]
    \centering
    \begin{tabular}{l|lccc}
        \hline
            \rule{0pt}{10pt}{Dataset}&{Method} & PSNR↑ & SSIM↑ & FID↓ \\
        \hline
            \multirow{4}{*}{Face} &
            \rule{0pt}{10pt}{Panini} & 27.13& 0.729& 14.15\\
            &\rule{0pt}{10pt}{Panini} w/ PD & \textbf{29.97} & \textbf{0.801}& \textbf{9.87}\\
            \cline{2-5} 
            &\rule{0pt}{10pt}{GLEAN} &27.20&0.74&12.63\\
            &\rule{0pt}{10pt}{GLEAN w/ PD} &\textbf{30.21}&\textbf{0.81}&\textbf{8.69}\\
        \hline
            \multirow{4}{*}{Cat} &
            \rule{0pt}{10pt}{Panini} &22.36&0.596&129.2\\
            &\rule{0pt}{10pt}{Panini w/ PD} &\textbf{23.52}&\textbf{0.623}&\textbf{118.9}\\
            \cline{2-5}
            &\rule{0pt}{10pt}{GLEAN} &22.74&0.588&62.92\\
            &\rule{0pt}{10pt}{GLEAN w/ PD} &\textbf{22.94}&\textbf{0.597}&\textbf{58.95}\\
        \hline
            \multirow{4}{*}{Church} &
            \rule{0pt}{10pt}{Panini} &19.27&0.483&\textbf{67.98}\\
            &\rule{0pt}{10pt}{Panini w/ PD} &\textbf{19.80}&\textbf{0.491}&69.20\\
            \cline{2-5}
            &\rule{0pt}{10pt}{GLEAN} &19.59&0.485&24.49\\
            &\rule{0pt}{10pt}{GLEAN w/ PD} &\textbf{19.99}&\textbf{0.500}&\textbf{24.03}\\
         \hline
    \end{tabular}
    \caption{\textbf{8$\times$ SR on different categories}. The use of PD significantly improves the PSNR, SSIM, and FID in most cases. It is worth noting that PD is parameter-free with negligible computational cost.}
    \label{tb1}
\end{table}

Likewise, we experiment 8$\times$ SR on LSUN cat and church datasets \cite{lsun} for GLEAN and Panini and their PD based version. Tab.~\ref{tb1} shows the quantitative results of 8$\times$ SR on three categories. As can be seen, the use of PD significantly improves all metrics on both Panini and GLEAN, implying the overall enhancement of \textit{consistency} and \textit{realness}. Fig.~\ref{curve}(a) shows the convergence curve on 8$\times$ face SR. We can observe a steady improvement of PSNR and FID during training.
Notably, ``Panini w/ PD" in 10K iterations achieves comparable PSNR and FID to Panini in 90K iterations, showing nine times of convergence acceleration.

To intuitively understand the eliminated inconsistencies by PD, we take the GT as references and visualize the error maps in Fig.~\ref{fig_8xsr}. Specifically, we subtract the result with GT to get the errors, then times five the absolute values and convert them into a color map. The results generated by GLEAN suffer color inconsistencies in most regions of the image. However, these inconsistencies are not a simple overall deviation of hue but are semantically related (e.g., it changes along the facial structures) and thus can not be eliminated by simple post-processing. However, with the use of PD, these low-frequency inconsistencies are gone, leaving only the high-frequency inconsistencies, which are acceptable considering SR as an ill-posed inverse problem.

\begin{figure}[t]
  \centering
  \includegraphics[width=1\linewidth]{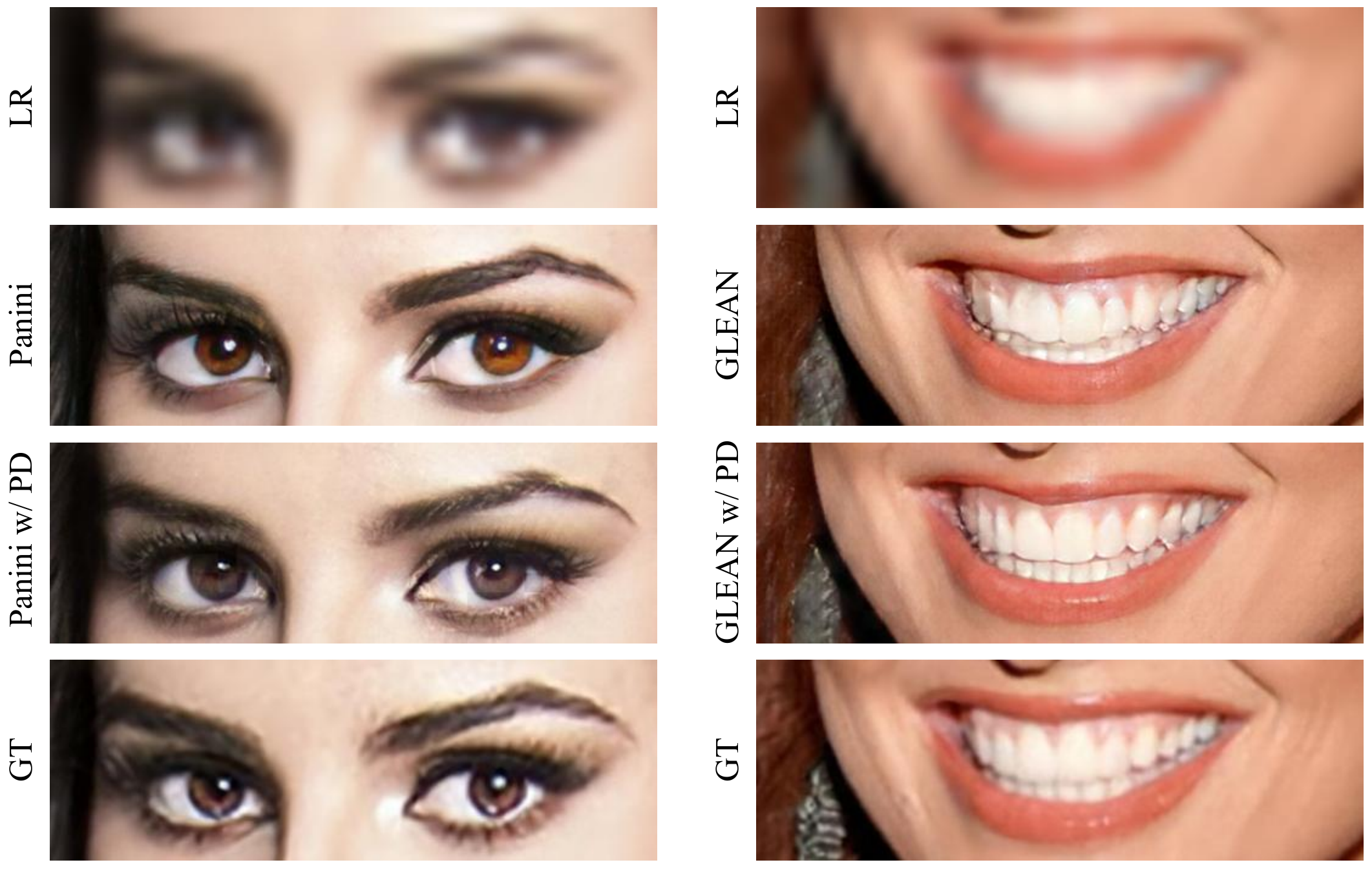}
  \caption{\textbf{Qualitative results on 16$\times$ face SR}. The use of PD can eliminate color deviation and reduce structural inconsistencies. }
\label{fig_16x} 
\end{figure}

\begin{table}[t]
    \centering
    \begin{tabular}{lcccc}

      \hline
         \rule{0pt}{10pt}{{Method}} &PSNR↑& SSIM↑ &MS-SSIM↑& FID↓ \\
         \hline
         \rule{0pt}{10pt}{PULSE} &21.68&0.676&0.596&42.71\\
         
         \rule{0pt}{10pt}{pSp} &18.91&0.680&0.526&39.88\\
         
         \rule{0pt}{10pt}{GFPGAN} &25.17&0.761&0.804&24.34\\
         \rule{0pt}{10pt}{GPEN} &26.07&\textbf{0.784}&0.820&31.89\\
         \rule{0pt}{10pt}{Panini} &27.18&0.758&0.843&\textbf{14.49}\\
         \rule{0pt}{10pt}{Panini w/ PD} &\textbf{27.81}&0.771&\textbf{0.851}&14.78\\
         \rule{0pt}{10pt}{GLEAN} &27.21&0.743&0.843&15.01\\
         \rule{0pt}{10pt}{GLEAN w/ PD} &27.69&0.754&0.848&15.27\\
         \hline
    \end{tabular}
    
\caption{\textbf{Comprehensive comparison on 16$\times$ face SR}. We compare Panini and GLEAN and their PD-based versions with state-of-the-art face SR methods. The involvement of PD significantly elevates all consistency metrics, i.e., PSNR, SSIM, and MS-SSIM. We attribute the slight rise of FID to the training stochasticity. Actually, the FID is comparable during training, as can be seen in Fig.~\ref{curve}.}    
\label{tb_16xsr}
\end{table}

\subsection{16$\times$ Super-Resolution}

The experiment on 16$\times$ SR is similar to the 8$\times$ SR experiment, except for the following changes: (1) We take the pretrained PULSE, pSp, GFPGAN, and GPEN, the state-of-the-art GAN Prior based face SR methods, for a comprehensive comparison. (2) The GAN loss weight are set as 1$\times10^{-3}$ for ``Panini w/ PD" and ``GLEAN w/ PD". (3) The training iteration is set as 300K.

Tab.~\ref{tb_16xsr} shows the quantitative comparisons on 16$\times$SR. We can see a significant improvement in consistency metrics with the use of PD. ``Panini w/ PD" gets the best score in PSNR and MS-SSIM and the second best score in SSIM and FID, achieving state-of-the-art in 16$\times$ face SR. Fig.~\ref{curve}(b) shows the convergence curve on 16$\times$SR. The use of PD elevates PSNR for all the evaluated iterations and yields comparable FID. Fig.~\ref{fig_16x} shows the qualitative comparison of Panini and GLEAN and their PD based versions. Panini can generate highly realistic details that are even superior to the GT. However, like most GAN prior based methods, it suffers inconsistencies, typically the eye colors. The use of PD helps eliminate such inconsistencies while maintaining superior realistic details. GLEAN may generate unreasonable tooth shapes. However, the use of PD can help yield more reasonable structures.

\begin{figure}[t]
  \centering
  \includegraphics[width=1\linewidth]{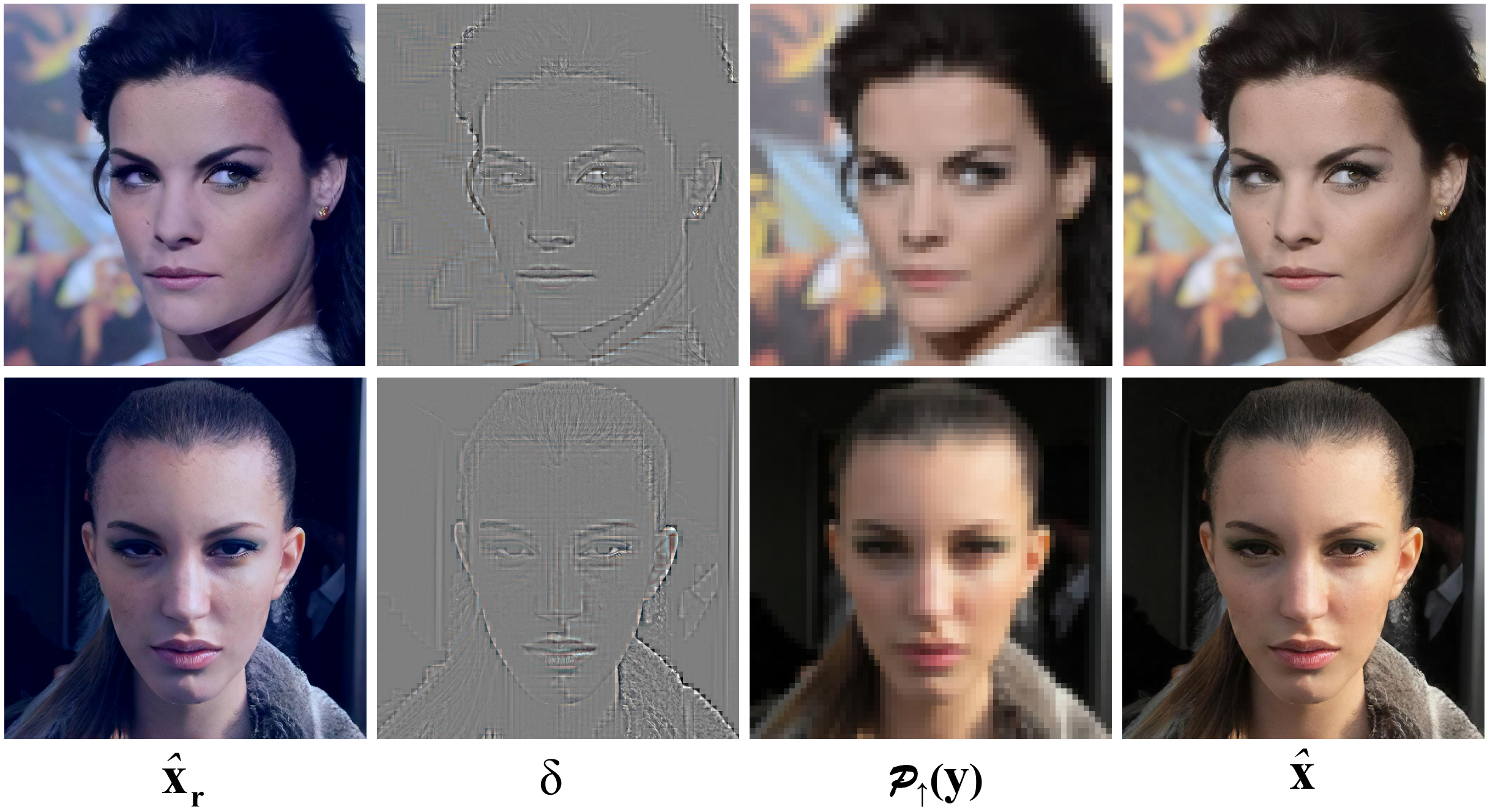}
  \caption{\textbf{Visualization of PD.} 
  $\hat{\mathbf{x}}_{r}$ represent the raw prediction of GAN prior network, $\delta$ is the high-frequency part of $\hat{\mathbf{x}}_{r}$. $\mathcal{P}_{\uparrow}(\mathbf{y})$ denotes the low-frequency contents inherited from LR image. The final result $\hat{\mathbf{x}}$ is yielded by adding $\mathcal{P}_{\uparrow}(\mathbf{y})$ with $\delta$. \textbf{(Zoom-in for the best view)}
  }
\label{ablation_delta} 
\end{figure}

\section{Ablation Studies}

\subsection{What Has the Network Learned?}
With the use of PD, we take the upsampled LR, i.e., $\mathcal{P}_{\uparrow}(\mathbf{y})$ as the low-frequency part, then only the high-frequency part of $\hat{\mathbf{x}}_{r}$ that GAN prior generates is needed to constitute the final HR result $\mathbf{y}$. Since we only extract high-frequencies from $\hat{\mathbf{x}}_{r}$, it does not necessarily own reasonable low-frequency contents. Interestingly, $\hat{\mathbf{x}}_{r}$ turns out to be reasonable in most low-frequency contents. Fig.~\ref{ablation_delta} visualizes $\hat{\mathbf{x}}_{r}$, high-frequency part $\delta$, and low-frequency part $\mathcal{P}_{\uparrow}(\mathbf{y})$. 

\subsection{What If We Do Not Use Pixel-Wise losses?}
The use of pixel-wise losses provides a clear learning target. However, it does not generalize well when the GT or LR suffers degradation. That is because the pixel-wise losses encourage learning the inverse process of the downsampling of GT. Thus when GT already contains degradation, the SR network trained on pixel-wise losses tends to replicate such degradation. 

Since PD can inherently assure \textit{consistency}, we may train the PD-based network without pixel-wise losses (e.g., $\ell_1$ and perceptual loss) and solely use adversarial losses for training. To stabilize the training, we take GLEAN as our backbone, with several critical changes: (1) we remove the ``decoder" since it is proved to be redundant \cite{glean_tpami}. (2) we do not predict the latent code from the ``encoder" and instead set the latent code z as random to encourage diversity, as is done in StyleGAN. (3) we apply PD to the network. This revised network is denoted as PDN.

We train PDN with the same training configurations mentioned in the experiment chapter, except for the absence of $\ell_1$ and perceptual loss. We find PDN works well on both 8$\times$ and 16$\times$ face SR tasks and shows superior robustness to GLEAN and Panini when facing unseen degradation. We evaluate PDN on two typical degradations.

\begin{figure}[t]
  \centering
  \includegraphics[width=1\linewidth]{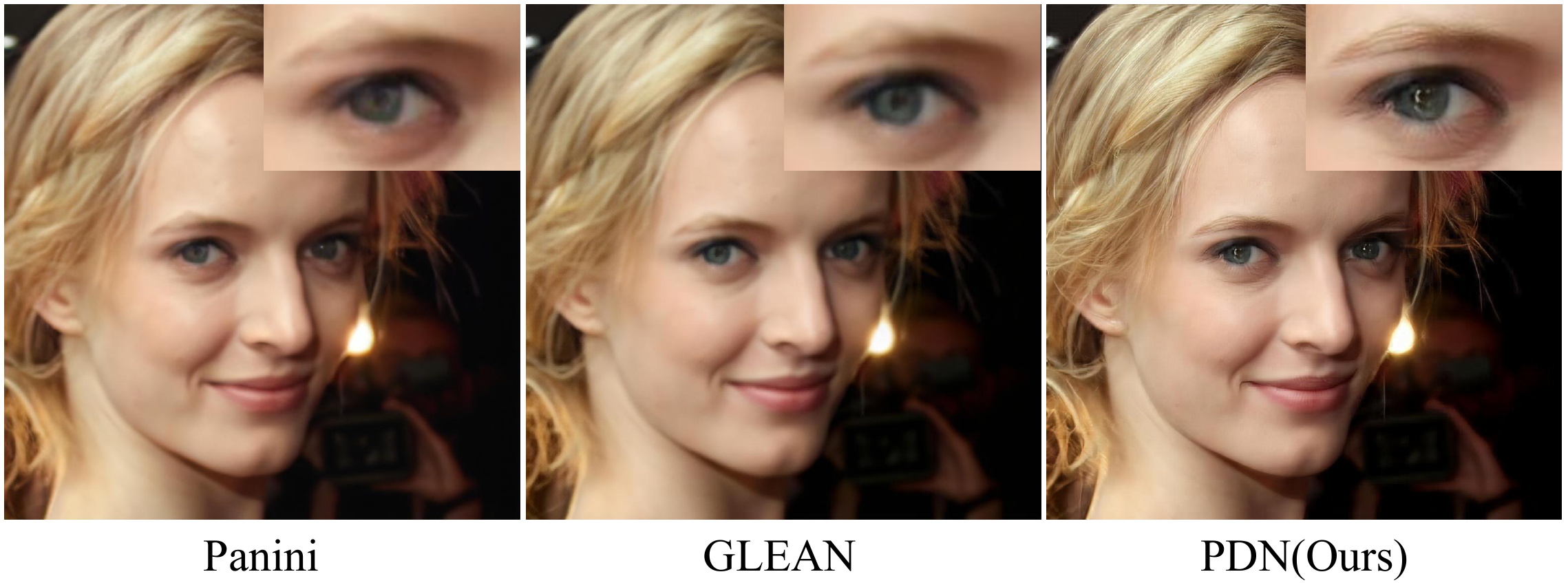}
  \caption{\textbf{Results on unseen downsamplings}. PDN yields clearer results when facing unseen downsamplings. Here the networks are all trained on 8$\times$ bicubic(alias) and tested on 8$\times$ bicubic(antialias).}
\label{fig_generalization} 
\end{figure}

\begin{table}[t]
    \centering
    \begin{tabular}{lcccc}

      \hline
         \multirow{2}{*}{{Method}} & \multicolumn{2}{c}{\rule{0pt}{10pt}{Bicubic}(antialias)}& \multicolumn{2}{c}{{Bilinear}(antialias)}\\
         &PSNR↑ & FID↓& PSNR↑ & FID↓\\
         \hline
         \rule{0pt}{10pt}{Panini} &29.40&30.47 &29.17 &37.09\\
         \rule{0pt}{10pt}{GLEAN} &30.13&24.26& 29.76& 29.50\\
         \rule{0pt}{10pt}{PDN(ours)} &\textbf{30.24}&\textbf{14.11}&\textbf{30.39}&\textbf{15.96}\\
         \hline
    \end{tabular}
    
    \caption{\textbf{Quantitative comparisons on unseen downsamplings}. We train PDN, Panini, and GLEAN on datasets that downsampled by 8$\times$ bicubic(alias) but test them on datasets that downsampled by 8$\times$ bicubic(antialias) and 8$\times$ bilinear(antialias) respectively. We can see that PDN achieves significantly better FID, indicating its robustness in restoring images from unseen downsamplings. Note that PDN is trained without using $\ell_1$, $\ell_2$, or perceptual losses.}
    \label{tb_generalization}
\end{table}

\begin{figure}[t]
  \centering
  \includegraphics[width=1\linewidth]{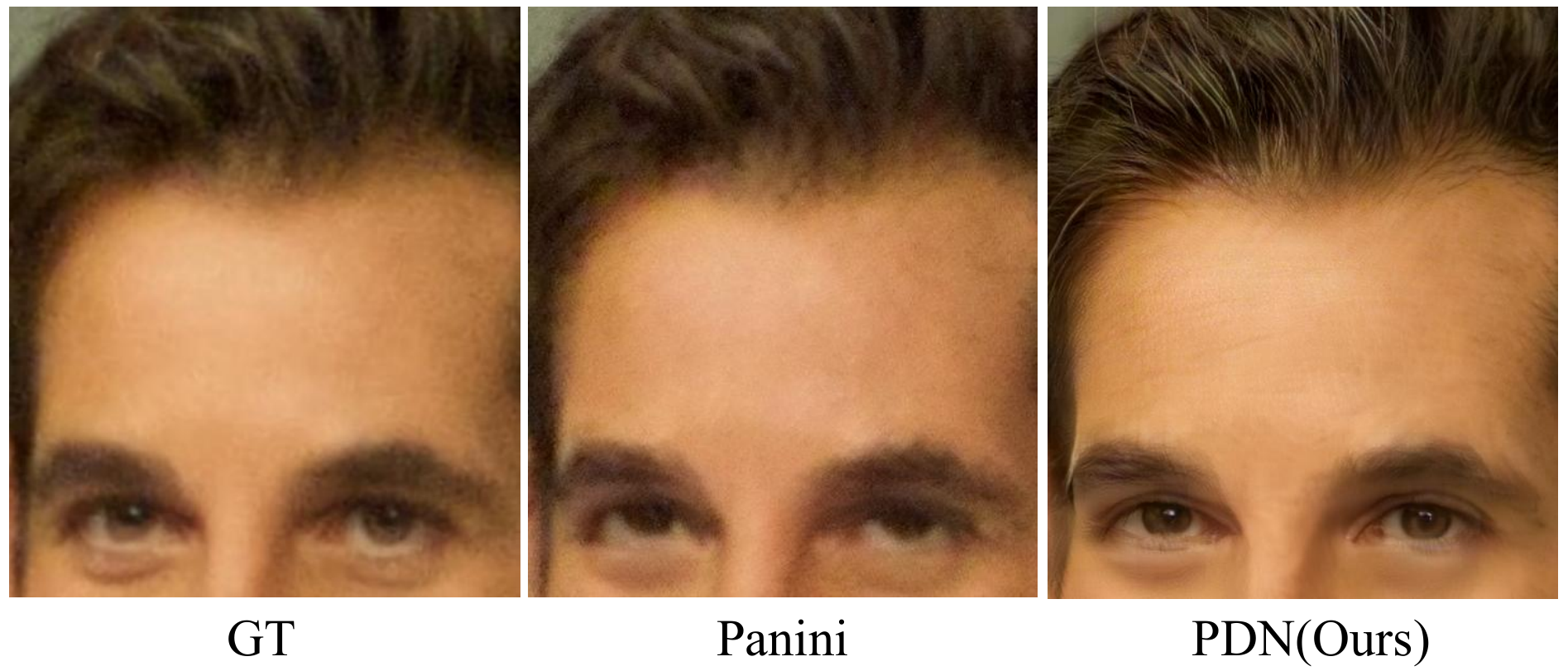}
  \caption{\textbf{Results on real-world degradation}. We can see that GLEAN tends to replicate the degradation that GT suffers, while PDN is not affected and tends to generate clear results. Note PDN only uses 16$\times$ bicubic(antialias) to synthesize LRs for training, without any simulated degradation.}
\label{fig_degradation} 
\end{figure}

\textbf{Unseen Downsamplings.} To evaluate the robustness of PDN on unseen downsampling methods, we generate training data by bicubic downsampling without antialiasing, noted as Bicubic(alias). In like manner, we generate Bicubic(antialias) and Bilinear(antialias) as testing data. We train PDN, GLEAN, and Panini on bicubic(alias) and evaluate them on Bicubic(antialias) and Bilinear(antialias). Note GLEAN and Panini are trained with $\ell_1$, perceptual and adversarial loss, while PDN is only with adversarial loss. Fig.~\ref{fig_generalization} shows the qualitative comparison. We can observe that PDN generates a clearer image than GLEAN and Panini. Tab.~\ref{tb_generalization} shows the PSNR and FID under Bicubic(antialias) and Bilinear(antialias). PDN yields much better FID than GLEAN and Panini, which is in accord with our visual observation. Besides, evaluations under other types of downsamplers like Box and Lanczos come to a similar conclusion.

\textbf{Real-World Degradation.} To evaluate the robustness of PDN on unknown real-world degradation, we cherry-pick low-quality images from CelebA-HQ dataset and use bicubic interpolation to downsample them as LR images. We use PDN, GLEAN, and Panini which are trained on 16$\times$SR. Not all three networks do not use any data augmentations to improve robustness during training. Fig.~\ref{fig_degradation} shows the results. We can see that the GT suffers unknown blur and noise, and Panini's SR result reproduces these degradations. However, the result generated by PDN seems not disturbed by the degradation, showing clear and realistic details. We also tried to remove PD from PDN and train it with pure adversarial loss but got much worse results, indicating that PD can stand as a powerful \textit{consistency} constraint to help the network converges to reasonable solutions.

PDN learns a more generalized pattern for SR, other than the inverse process of the downsampling. However, PDN is no better than state-of-the-art SR methods when dealing with LR that is downsampled from high-quality GT. We believe that is because PDN lacks detailed guidance. We hope this guide can be solved in the future without resorting to pixel-wise loss.

\section{Conclusions and Discussions}
This paper presents a novel method to eliminate inconsistencies for GAN prior based super-resolution networks. As is shown in experiments, our proposed method can be easily applied to different backbones, accelerating their training convergence and yielding better \textit{consistency}. Our method also shows potential in dealing with unseen downsamplings or real-world degradation.    

There are a few limitations of our method. Firstly, our method is under an ideal assumption that the LR is clean. However, if the degradation that LR suffers damage the low-frequency contents, PD will pass the damaged low-frequencies directly to the SR result. A simple solution is to pre-process the degraded LR into a clean one. Secondly, PD may result in evident block artifacts at the early stages of training, but as the training goes on, those artifacts become imperceptible (see if you can observe block artifacts in Fig.~\ref{fig_16x}). Thirdly, PD is sensitive to hyperparameter selection. See Appendix B for analysis.

\bibliography{aaai23}

\clearpage
\section{Appendix}
\subsection{A: Details of Training}
For fair evaluation, the training of Panini and ``Panini w/ PD" follow the official setting of Panini while the training of GLEAN and ``GLEAN w/ PD" follow the official setting of GLEAN \cite{panini,glean}. Specifically, a pretrained StyleGAN2 generator \cite{stylegan2} is adopted as the GAN Prior module and the corresponding StyleGAN2 discriminator is used for adversarial training. For GLEAN and ``GLEAN w/ PD", the parameters of StyleGAN2 are fixed during training. For Panini and ``Panini w/ PD", all parameters are set as trainable. 

Let $\mathcal{D}_{\uparrow}(\cdot)$ denotes the forward process of the network (Panini or GLEAN). Then the SR result of Panini or GLEAN can be written as:
\begin{equation}
    \hat{\mathbf{x}} = \mathcal{D}_{\uparrow}(\mathbf{y})   
\end{equation}
Instead, the SR result of ``Panini w/ PD" and ``GLEAN w/ PD" is yielded through PD:
\begin{equation}
    \hat{\mathbf{x}} = \mathcal{P}_{\uparrow}(\mathbf{y}) + \hat{\mathbf{x}}_{r} - \mathcal{P}_{\uparrow}(\mathcal{P}_{\downarrow}(\mathcal{D}_{\uparrow}(\mathbf{y})))   
\end{equation}
The training objective consists of three parts: $\ell_1$ or $\ell_2$ loss, perceptual loss, and adversarial loss.

The standard $\ell_1$ or $\ell_2$ loss is used to minimize the errors between the GT image $\mathbf{x}$ and the SR result $\hat{\mathbf{x}}$:
\begin{equation}
    \mathcal{L}_{1}= \left\| \hat{\mathbf{x}} - \mathbf{x} \right\|_{1}, \quad \mathcal{L}_{2}= \left\| \hat{\mathbf{x}} - \mathbf{x} \right\|_{2}.   
\end{equation}
The VGG perceptual loss \cite{perceptualloss} is used to provide better guidance for reconstruction:
\begin{equation}
    \mathcal{L}_{vgg}= \left\| \mathcal{H}_{vgg}(\hat{\mathbf{x}}) - \mathcal{H}_{vgg}({\mathbf{x}}) \right\|_{2},    
\end{equation}
where $\mathcal{H}_{vgg}(\cdot)$ denotes the intermediate feature maps embedded in the VGG16 network. 

Further, we adopt vanilla adversarial loss \cite{gan} to enhance the visual quality:
\begin{equation}
    \mathcal{L}_{gen}= \log(1 - \mathcal{H}_{dis}(\hat{\mathbf{x}})),    
\end{equation}
where $\mathcal{H}_{dis}(\cdot)$ denotes the discriminator of the aforementioned pretrained StyleGAN2. 

The complete objective functions can be formulated as $\mathcal{L}_{G}$ and $\mathcal{L}_{D}$ respectively. For Panini and ``Panini w/ PD", they use $\ell_1$ loss:
\begin{equation}
    \mathcal{L}_{G}= \lambda\mathcal{L}_{1}+\alpha\mathcal{L}_{vgg}+\beta\mathcal{L}_{gen},    
\end{equation}
\begin{equation}
    \mathcal{L}_{D}= \log(\mathcal{H}_{dis}(\mathbf{x})) + \log(1-\mathcal{H}_{dis}(\hat{\mathbf{x}})).   
\end{equation}
while for GLEAN and ``GLEAN w/ PD", they use $\ell_2$ loss:
\begin{equation}
    \mathcal{L}_{G}= \lambda\mathcal{L}_{2}+\alpha\mathcal{L}_{vgg}+\beta\mathcal{L}_{gen},    
\end{equation}
\begin{equation}
    \mathcal{L}_{D}= \log(\mathcal{H}_{dis}(\mathbf{x})) + \log(1-\mathcal{H}_{dis}(\hat{\mathbf{x}})).   
\end{equation}
During training, we minimize $\mathcal{L}_{G}$ and maximize $\mathcal{L}_{D}$ iteratively. The training iteration is set as 100K for 8$\times$ SR and 300K for 16$\times$ SR. For 16$\times$ SR, hyperparameters $\lambda$, $\alpha$, $\beta$ are set as 1, $10^{-2}$, $10^{-2}$, respectively. For 16$\times$ SR, $\beta$ is set to $10^{-3}$ to stabilize the training. See section B for analysis.

\subsection{D: Extending to Other Tasks}
The way of construction pseudo-inverse in PD can be easily extended to other image restoration (IR) tasks, thereby yielding a series of practical operators for null-space learning on different IR tasks. For example, colorization and compressed sensing. We provide PyTorch-like implementation of these hand-designed operator and their pseudo-inverse here.

\lstset{ %
language=python,                
basicstyle=\scriptsize,           
numbers=left,                   
numberstyle=\tiny\color{gray},  
stepnumber=2,                   
numbersep=5pt,                  
backgroundcolor=\color{white},      
showspaces=false,               
showstringspaces=false,         
showtabs=false,                 
frame=single,                   
rulecolor=\color{black},        
tabsize=2,                      
captionpos=b,                   
breaklines=true,                
breakatwhitespace=false,        
title=\lstname,                 
keywordstyle=\color{blue},          
commentstyle=\color{green},       
stringstyle=\color{orange},         
escapeinside={\%*}{*)},            
morekeywords={*,...}               
}
\vspace{0.3cm}
\begin{lstlisting}

def color2gray(x):
    coef=1/3
    x = x[:,0,:,:]*coef + x[:,1,:,:]*coef + x[:,2,:,:]*coef
    return x.repeat(1,3,1,1)

def gray2color(x):
    coef=1/3
    x = x[:,0,:,:]
    base = 1.
    return torch.stack((x*coef/base, x*coef/base, x*coef/base), 1)    
    
def PatchUpsample(x, scale):
    n, c, h, w = x.shape
    x = torch.zeros(n,c,h,scale,w,scale) + x.view(n,c,h,1,w,1)
    return x.view(n,c,scale*h,scale*w)

# Implementation of A and its pseudo-inverse Ap    

H, W = 1024, 1024  # image height & width
    
if IR_mode=="colorization":
    A = color2gray
    Ap = gray2color
    
elif IR_mode=='compressed sensing':
    C = 3  # channel number
    B = 32  # block size
    N = B * B
    q = int(torch.ceil(torch.Tensor([cs_ratio * N])))
    U, S, V = torch.linalg.svd(torch.randn(N, N))
    A_weight = U.mm(V)  # CS sampling matrix weight
    A_weight = A_weight.view(N, 1, B, B)[:q]
    A = lambda z: torch.nn.functional.conv2d(z.view(batch_size * C, 1, H, W), A_weight, stride=B)
    Ap = lambda z: torch.nn.functional.conv_transpose2d(z, A_weight, stride=B).view(batch_size, C, H, W)
      
elif IR_mode=="super resolution":
    A = torch.nn.AdaptiveAvgPool2d((H//scale,W//scale))
    Ap = lambda z: PatchUpsample(z, scale)
\end{lstlisting}

\begin{figure*}[t]
  \centering
  \includegraphics[width=1\linewidth]{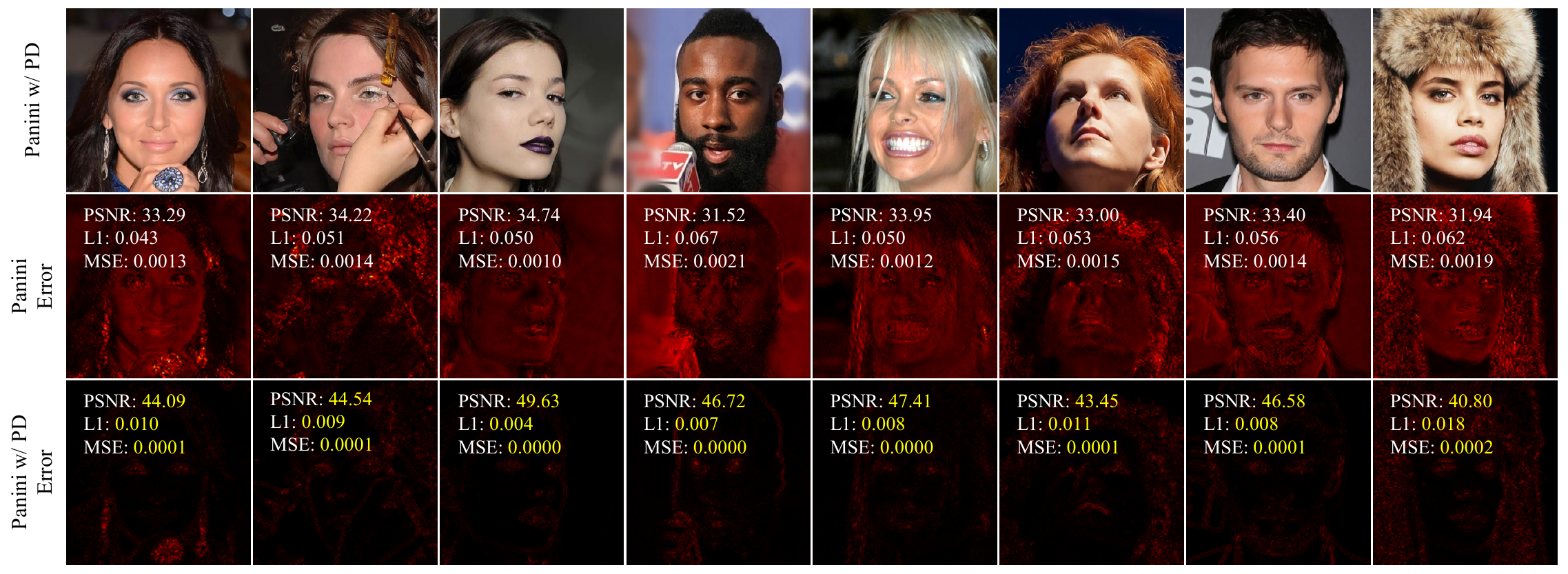}
  \caption{\textbf{Comparison on \textit{consistency}}. We randomly pick eight SR results and visualize the \textit{consistency} error, defined by $\mathcal{P}_{\downarrow}(\mathbf{\hat{x}})-\mathbf{y}$, where $\mathbf{y}$, $\mathbf{\hat{x}}$, $\mathcal{P}_{\downarrow}$ denote the LR image, SR result and average-pooling, respectively. The first row shows the SR results of ``Panini w/ PD". The second and third rows show the \textit{consistency} error maps of Panini and ``Panini w/ PD" (we times five the error values to enhance the visual salience of the error maps). We also calculate the PSNR, L1 error and MSE of each $(\mathcal{P}_{\downarrow}(\mathbf{\hat{x}}),\mathbf{y})$ pairs for better comparison. It is clear that the inconsistencies are eliminated by PD. It is worth noting that the remaining errors of ``Panini w/ PD" are caused by the threshold truncation when transforming the image format.}
\label{fig:consistency} 
\end{figure*}

\newpage
\subsection{B: Hyperparameter Selection}
Our SR results may be inconsistent with LR results due to threshold truncation when converting tensors into image format. For a instance, given $\mathbf{y}=(0)$ and $\hat{\mathbf{x}}_{r}=(0, 1)$, we can yield $\hat{\mathbf{x}}=(-0.5, 0.5)$ through PD. Now the average value of $\hat{\mathbf{x}}$ equals to $\mathbf{y}$. However, once we convert $\hat{\mathbf{x}}$ into image format, e.g., within range of [0,1], it becomes $\hat{\mathbf{x}}=(0, 0.5)$ after threshold truncation and $\hat{\mathbf{x}}$ is no longer consistent to $\mathbf{y}$. Due to the same reason, PD is sensitive to hyperparameter selection. When facing more challenging SR scales, e.g., 16$\times$ or higher, it usually needs to enhance the guidance of $\ell_1$ or $\ell_2$ loss to stabilize the training, e.g., elevating the weight of $\ell_1$ or $\ell_2$ loss or decreasing the GAN loss weight.

\begin{figure}
  \centering
  \includegraphics[width=1\linewidth]{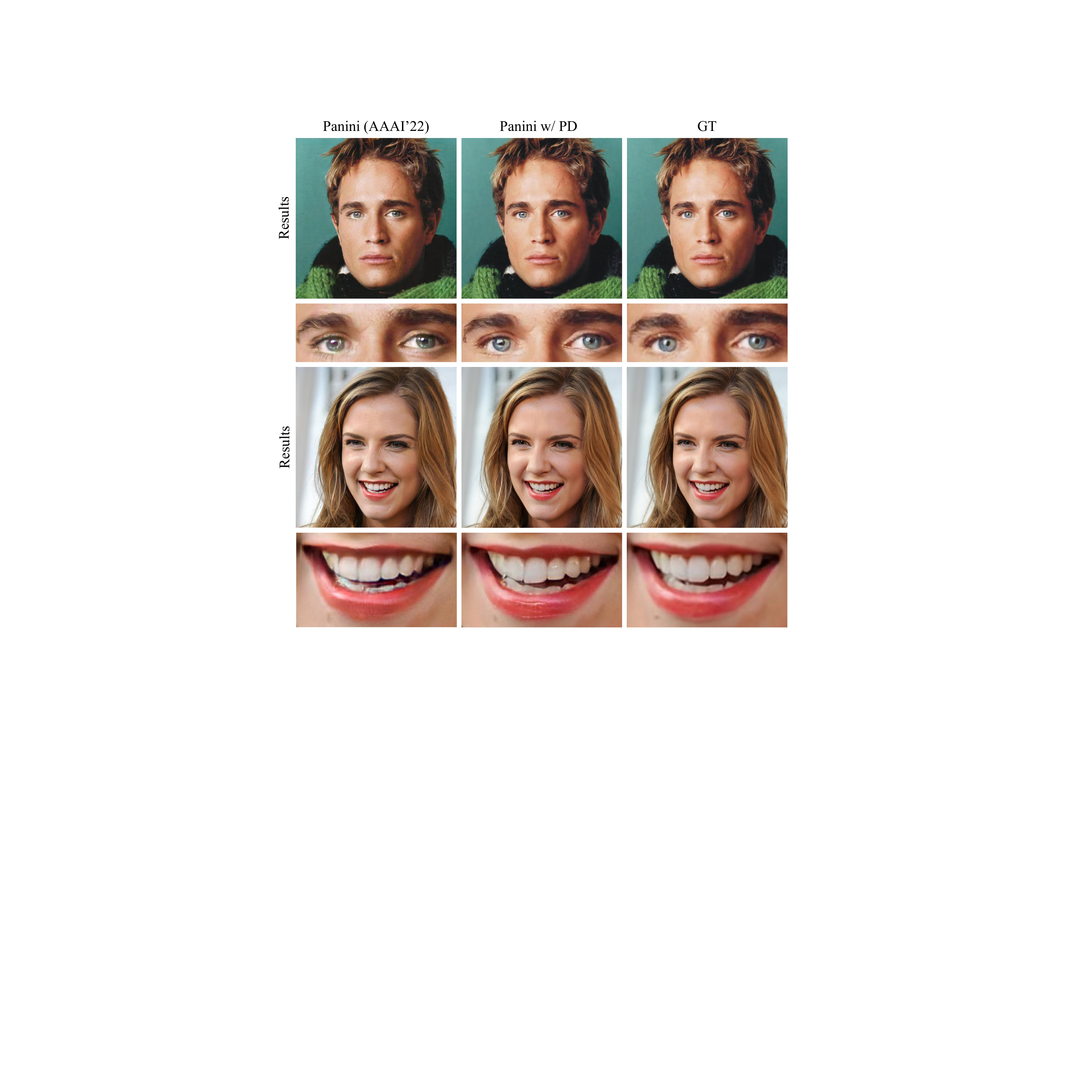}
  \caption{\textbf{PD helps maintain consistency in the early training stages} and thus accelerates the training convergence. Here the networks are all at 50k iterations of training (half of the whole training iterations). We can see that ``Panini w/ PD" performs better than Panini in both realness and consistency, revealing its faster convergence speed.}
\label{fig:ap4} 
\end{figure}

\subsection{C: Additional Results}
We visualize the intermediate result during training, providing a intuitive feeling for acceleration of convergence that PD brings, as is shown in Fig.~\ref{fig:ap4}.

\begin{figure*}
  \centering
  \includegraphics[width=1\linewidth]{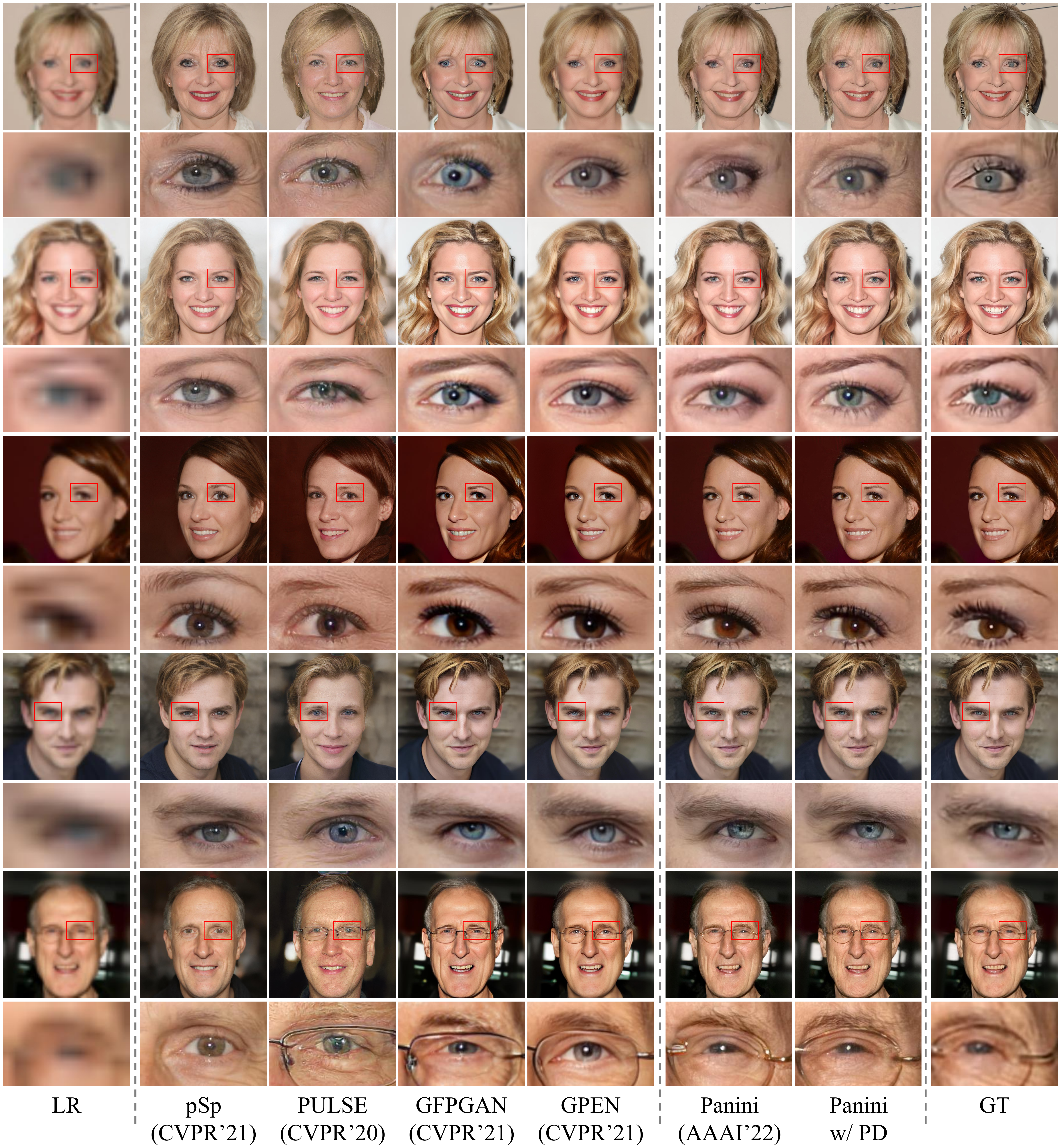}
  \caption{\textbf{Comprehensive comparison on 16$\times$ face SR}. The use of PD helps eliminate color and structural inconsistencies in Panini and refreshes state-of-the-art performance in 16$\times$ face SR task.}
\label{fig:subjective} 
\end{figure*}

\begin{figure*}
  \centering
  \includegraphics[width=1\linewidth]{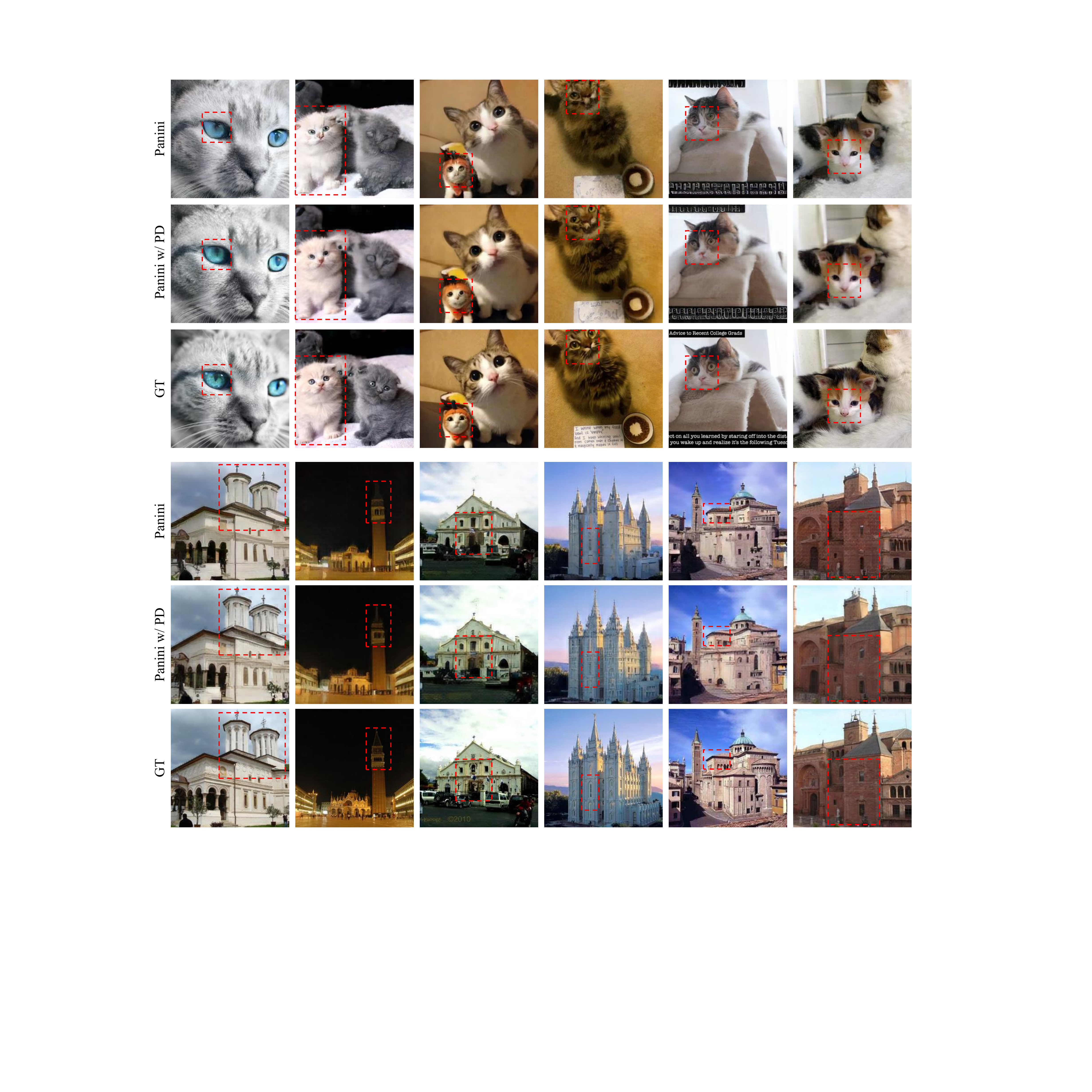}
  \caption{\textbf{Qualitative results of 8$\times$ SR on LSUN cat and church}. PD helps correct the color distributions and yield more reasonable structures.}
\label{fig:cat} 
\end{figure*}

Fig.~\ref{fig:consistency} visualizes the \textit{consistency} error. Fig.~\ref{fig:subjective} compares the visual results on 16$\times$ face SR task, including PULSE \cite{pulse}, pSp \cite{psp}, GFPGAN \cite{gfpgan}, GPEN \cite{gpen}, Panini \cite{panini}, and ``Panini w/ PD". The use of PD helps eliminate color and structural inconsistencies in Panini and refreshes state-of-the-art performance in 16$\times$ face SR task. Fig.~\ref{fig:cat} compares the visual results of 8$\times$ SR on LSUN cat and church datasets.

\end{document}